\documentclass[english]{article}
\usepackage[T1]{fontenc}
\usepackage[latin9]{inputenc}
\usepackage{babel}
\usepackage{amsmath}
\usepackage{amsthm}
\usepackage{amssymb}
\usepackage{graphicx}
\usepackage{microtype}
\usepackage[unicode=true,
 bookmarks=false,
 breaklinks=false,pdfborder={0 0 1},backref=section,colorlinks=false]
 {hyperref}
\hypersetup{
 hidelinks}

\makeatletter

\usepackage{babel}

\usepackage{ijcai22}

\usepackage{times}
\usepackage{soul}
\usepackage{url}
\usepackage[small]{caption}
\usepackage{amsthm}
\usepackage{booktabs}
\usepackage{algorithm}
\usepackage{algorithmic}
\usepackage{microtype}

\title{IJCAI--22 Formatting Instructions}

\author{
Minh-Tri Nguyen
\and
Thin Nguyen\And
Truyen Tran
\affiliations
Applied Artificial Intelligence Institute, Deakin University, Australia
\emails
\{minhtri,thin.nguyen,truyen.tran\}@deakin.edu.au
}

\makeatother

\begin{document}
\title{Learning to Discover Medicines}
\maketitle
\begin{abstract}
Discovering new medicines is the hallmark of human endeavor to live
a better and longer life. Yet the pace of discovery has slowed down
as we need to venture into more wildly unexplored biomedical space
to find one that matches today's high standard. Modern AI--enabled
by powerful computing, large biomedical databases, and breakthroughs
in deep learning--offers a new hope to break this loop as AI is rapidly
maturing, ready to make a huge impact in the area. In this paper we
review recent advances in AI methodologies that aim to crack this
challenge. We organize the vast and rapidly growing literature of
AI for drug discovery into three relatively stable sub-areas: (a)
\emph{representation learning} over molecular sequences and geometric
graphs; (b) \emph{data-driven reasoning} where we predict molecular
properties and their binding, optimize existing compounds, generate
\emph{de novo} molecules, and plan the synthesis of target molecules;
and (c) \emph{knowledge-based reasoning} where we discuss the construction
and reasoning over biomedical knowledge graphs. We will also identify
open challenges and chart possible research directions for the years
to come. 
 
\end{abstract}

\section{Introduction}

\label{sec:intro} The COVID-19 pandemic has triggered an unprecedented rise of investment
capital in AI for drug discovery (DD), the process of identifying
new medicines for a druggable target \cite{Zhang2021TheReport}. Recent
breakthroughs in AI present a great opportunity to break the so-called
\emph{Eroom's Law} in DD--the inverse of the well-known Moore's Law--dictating
that the rate of FDA drug approval is slowing down despite a huge
increase in development cost \cite{Scannell2012DiagnosingEfficiency}.
Enabled by deep learning advances, powerful computing, and large databases,
modern AI is ready to make a huge impact through \emph{in silico}
processes to supplement and sometimes replace the \emph{in vitro}
counterparts of drug development. For a wide range of problems, from
determining the 3D structure of proteins to predicting drug-target
binding, to generating synthesizable molecules, AI has helped change
the DD landscape in recent years. The reverse also holds: The problems
in DD necessitate new advances in AI methodologies to deal with the
new scope and complexity typically not seen in traditional application
domains of AI such as computer vision and language processing.

\emph{There are three major DD questions AI can help answer}. The
first is, given the molecule, what are its chemo-biological and therapeutic
properties? Second, for a given target, what kind of molecules will
therapeutically modify its functions? Finally, given a molecule, how
can we synthesize and optimize the molecule from the available compounds,
meaning solving the problems of synthetic tractability and reaction
planning?

\begin{figure*}[h]
\centering{}\includegraphics[width=0.8\textwidth]{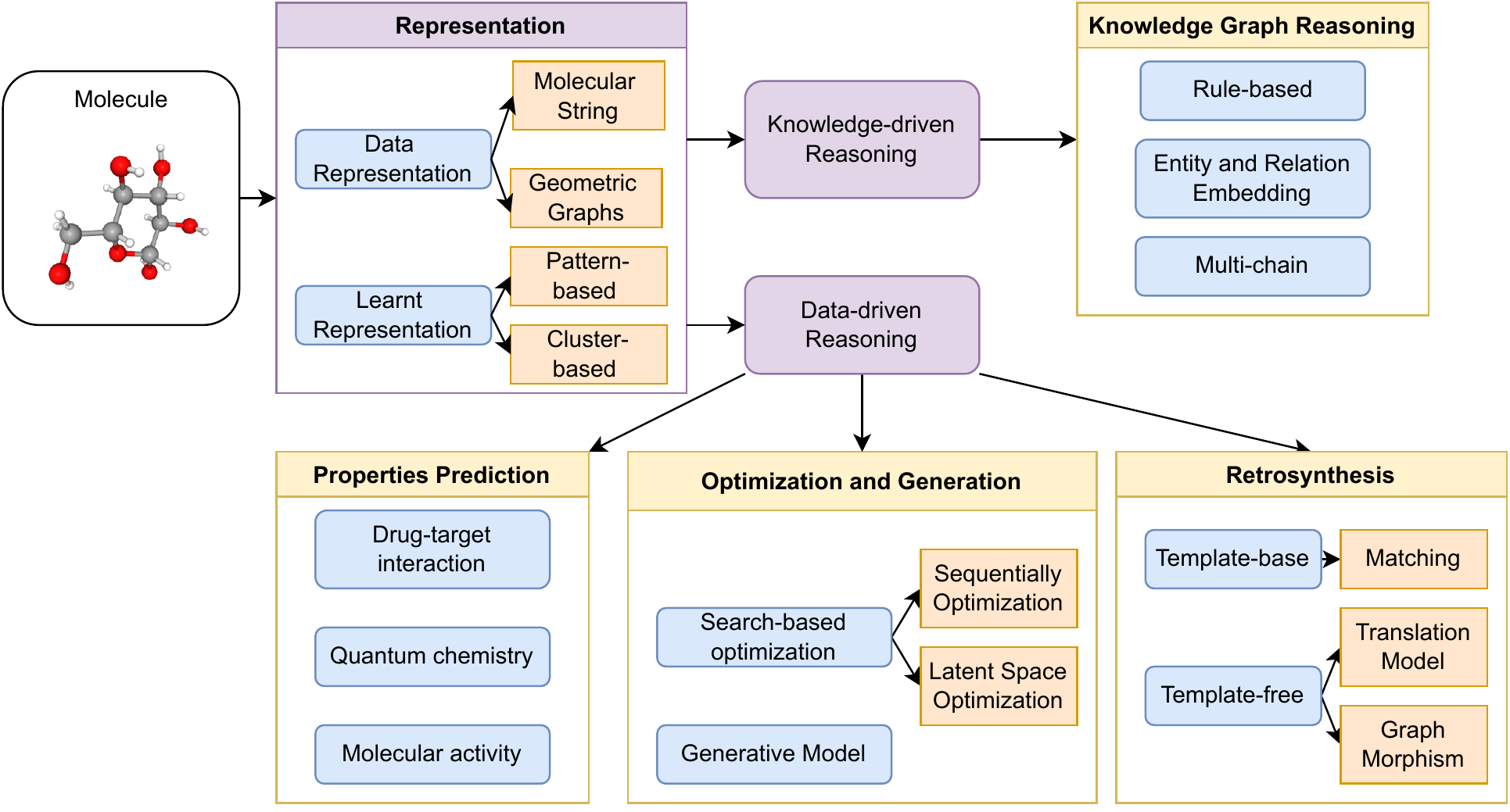}
\caption{Three aspects of AI in drug discovery: (i) Transforming the biological
data into representations readable by computer; (ii) data-driven reasoning
in which models estimated from data are used to infer properties,
optimize, generate molecules and plan synthesis; and (iii) reasoning
with biomedical knowledge graphs. \label{fig:taxonomy}}
\end{figure*}

In this survey, we bring in the machine learning (ML) and reasoning
perspectives for answering these questions, with an emphasis on recent
developments. Each question poses representation, learning, and reasoning
sub-problems. This is because drugs, targets, and the hosting environments
need to be represented in machine comprehensible formats. Modern ML
suggests that the representations should be learned instead of handcrafted
to best explore the power of computing and the richness of data (Sec.~\ref{sec:Learning-Representations}).
Once learning has been completed, the next phases of prediction, search
and discovery are performed using reasoning methods that leverage
the learned models (Sec.~\ref{sec:Data-driven-Reasoning}) and the
vast domain knowledge (Sec.~\ref{sec:Knowledge-based-Reasoning}).
Before concluding, we will discuss the remaining challenges and opportunities
for AI/ML in this important area (Sec.~\ref{sec:Challenges-and-Opportunities}).
See Fig.~\ref{fig:taxonomy} for a taxonomy of the problem space,
which we follow in the paper.

\section{Learning Representations \label{sec:Learning-Representations}}

The first step of applying AI/ML is forming a computer-readable representation
of biomedical entities and concepts. We will primarily focus on the
drug-target pairs. A \emph{drug} is a small molecule while a \emph{target}
such as protein is a large (macro) one. Typically, in drugs, we are
concerned with atoms and bonds, while in protein, we are concerned
with amino acids.

\subsection{Representing data}

\subsubsection{Molecular strings}

The atoms and bonds of a small molecule can be efficiently represented
as a string of ASCII characters. A popular representation is SMILES
(Simplified Molecular-Input Line-Entry System), which can be decoded
back to the atom graph. However, the SMILES string may not be unique
as a molecule can have different SMILES forms. 

Likewise, a protein can also be represented by a string of characters
varying in length. Each character represents one of the 20 amino acids.
We often incorporate the evolutionary information of the target sequence
by searching for related proteins to form multiple sequence alignment
(MSA) and extracting evolutionary information (EI). The intuition
is that the important substructure of protein remains stable through
evolution. The EI has shown its effectiveness in several tasks such
as protein folding prediction \cite{Jumper2021HighlyAlphaFold}. 

\subsubsection{Geometric graphs}

\label{sec:graphrep} A richer and more precise representation of
a molecule is attributed graph. A molecular graph is defined as $G=(V,E)$
where $V$ is the atom set of the molecule and $E$ is the edge set
of bondings between atoms. To balance between 3D structural information
and simplicity, 2D representation via attributed graph can be used.
For example, in the case of protein, we predict the distance/contact
between residues to form the contact/distance map. The contact/distance
map is then used as an adjacency matrix of an attributed graph where
each node represents a residue and edges represent the contact/distance
between residues. 

Indeed, many biomedical problems are well cast into graph reasoning:
Molecule properties prediction as \emph{graph classification/regression},
drug-target binding as \emph{graph-in-graph}, chemical-chemical interaction
as \emph{graph-graph pairing}, molecular optimization as \emph{graph
edit/translation}, and finally chemical reaction as \emph{graph morphism}.

\subsubsection{Biochemical descriptors}

For small molecules, \emph{fingerprints} are often used to encode
the 2D structure into a vector. One approach is using \emph{structural
keys} to encode the structure of the molecule into a bit string, each
bit represents the presence or absence of predefined feature such
as substructure or fragment \cite{Durant2002ReoptimizationDiscovery}.
The structural keys suffer from the lack of generalization because
it depends on the pre-defined fragments and substructure to encode
the molecule. The alternative approach is \emph{hashed fingerprints}
which encode the counting of molecular fragments into numeric values
using a hash function, without relying on a pre-defined library. Based
on the fragment enumerating process, the hashed fingerprints can be
categorized into path-based and circular types \cite{Rogers2010Extended-ConnectivityFingerprints}.
In \cite{Duvenaud2015ConvolutionalFingerprints}, manual fingerprint
extraction is replaced by a \emph{learnable} hashing function on the
convolution over the molecular graph.

For proteins, a set of descriptors are constructed based on the amino
acids, their appearance frequency \cite{Gao2005PredictionSequence},
their individuals \cite{Sandberg1998NewAcids} or autocorrelation
\cite{Feng2000PredictionAcids} physical-chemical characteristic,
and sequence-order feature derived from physicochemical distance.
With the advance in the protein structure prediction, the distance
between residues is also considered in protein description \cite{Xu2020DeepEngineering}.

\subsection{Learnt representation }

The string representation of molecules makes it ready for applying
language modeling techniques, assuming the existence of statistical
sequential patterns, akin to those found in linguistic grammars. Because
the atom structure in the molecule follows a set of rules such as
valance, we can view it as grammar rules in chemical language. This
is not limited to string representation and can be extended to other
types of representation such as graphs. Several unsupervised representation
learning models have exploited the structural patterns to learn the
molecule representations. An early work in sequential representation
learning, word2vec \cite{Mikolov2013EfficientSpace}, learns the representation
by using the predicting neighbour tokens as a self-supervised task.
Recently, Transformers with BERT-like masking strategy has become
a popular technique to learn sequential representations, e.g., out
of molecular SMILES sequence \cite{Chithrananda2020ChemBERTa:Prediction}.
Likewise, the graph representation of molecules allows us to learn
the graph structure patterns. For example, DeepWalk \cite{Perozzi2014DeepWalk:Representations}
learns the node representation of the given graph structure using
random walks to learn the pattern of nodes' neighbour. Grover \cite{Rong2020Self-SupervisedData}
uses subgraph masking as contextual properties prediction and graph
motif prediction.

An increasingly popular strategy is through exploring the local structure
in the chemical space by using \emph{contrastive learning} for estimating
representations. This works by minimizing an energy-based loss to
keep the distance in the embedding space small for similar molecules
and large for dissimilar pairs \cite{Qiu2020GCC:Pre-Training}. The
main difference between these contrastive losses is the number of
positive, negative samples and how the pairs are sampled.

\section{Data-driven Reasoning \label{sec:Data-driven-Reasoning}}

There are three basic questions in drug discovery. The first question
is determining \textbf{whether the given molecule is drug-like}, meaning
having therapeutic effects on druggable targets. The second question
is given a biological target, \textbf{what are the candidate compounds}
that are likely to modify their functions in a desired way? This question
addresses two sub-problems: searching and generating. The former is
about finding a suitable molecule from an approved list, and the latter
is generating a \emph{de novo} molecule tailored to the target. The
final question is given a molecule, \textbf{how can we make it}? This
question addresses the sub-problems of synthetic tractability, reaction
planning, and retrosynthesis.

\subsection{Molecular property prediction}

\label{sec:molprop}

The first question is predicting the molecule's properties. There
are a wide range of prediction tasks, from predicting the drug-likeness,
which targets it can modulate, to predicting molecule dynamics/kinetics/effects/metabolism
if administered orally or via injection. Molecule property prediction
is a fundamental task in many stages of drug discovery.

The task is a specific case of many-body systems where we predict
the emergent properties of a group of interacting objects. The most
popular first-principle technique to tackle this general problem is
Density Functional Theory (DFT) to approximate the wave function,
which describes the quantum state of an isolated quantum system in
the many-body system. However, calculating DFT is computationally
expensive and can take up to $\mathcal{O}(10^{3})$ seconds for a
medium-sized molecule, making rapid screening over millions of potential
candidates intractable. 

A recent approach is to approximate DFT calculations by learning a
graph neural network (GNN) over the molecular graphs, which can be
trained on a large pre-computed dataset. Once trained GNNs can run
many orders of magnitudes faster than precise DFT methods with reasonable
accuracy. A popular type of GNNs is the Message Passing Neural Network
(MPNN) \cite{Gilmer2017NeuralChemistry,Unke2021SpookyNet:Effects}
which models the atoms interaction with message passing function,
update function, and readout function. MPNN and its cousin, the Graph
Convolutional Network (GCN), have since been frequently used in predicting
physical chemistry properties (e.g., water solubility, hydrophobicity),
physiology (e.g., toxicity), and biophysics (bind affinity) \cite{Altae-Tran2017LowLearning,Yang2019AnalyzingPrediction}.
Alternatively, the molecule properties prediction can also be formulated
as a reasoning process to answer a query (of a specific property),
and thus lends itself to more elaborate neural networks such as Graph
Memory Networks (GMN)  \cite{Pham2018GraphPrediction}.

\subsection{Drug-target affinity prediction}

\label{sec:protein_drug_binding}

Drug-target binding affinity indicates the strength of the binding
force between the target protein and its ligand (drug or inhibitor)
\cite{Ma2018OverviewInteraction}. There are two main approaches:
the structural approach and the non-structural approach \cite{Thafar2019ComparisonAffinities}.
Structural methods utilize the 3D structure of proteins and ligands
to run the interaction simulation between proteins and ligands. On
the other hand, the non-structural approach relies on ligand and protein
features such as sequence, hydrophobic, similarity and other structural
information.

\paragraph{Structural approach}

The structure-based approach usually relies on molecular docking which
simulates the post-binding 3D conformation of drug-target complex.
As there are several possible conformations, the simulated structure
is evaluated using a scoring function. The scoring function can vary
from the molecular mechanics' interaction energies \cite{Meng1992AutomatedEvaluation}
to machine learning predicted value derived from protein and drug
features \cite{Kundu2018AProperties,Gomes2017AtomicAffinity}, or
3D convolution on 3D structure \cite{Stepniewska-Dziubinska2018DevelopmentPrediction}.

\paragraph{Non-structural approach}

The non-structural approach relies on the drug/target similarity,
and structural features such as protein sequence or secondary structure
without relying on calculating the exact 3D structure of the drug-target
complex. Popular among them are kernel-based methods which employ
kernel functions to measure the molecule similarity \cite{Cichonska2017Computational-experimentalInhibitors}.
Alternatively, we can use drug-drug, target-target, and drug-target
similarity features as input for a classifier/regressor \cite{He2017SimBoost:Machines}.
More recently neural networks have become common as they can learn
the drug and target representations instead of handcrafting them.
For sequences, 1D convolution \cite{Ozturk2018DeepDTA:Prediction},
BiLSTM \cite{Zheng2020PredictingSystem}, or language model feature
\cite{Nguyen2021GEFA:Prediction} are used to encode the biological
sequence to the latent space. The drawback of sequential features
is that they ignore the structural information of the drug and the
target which also plays a critical role in the drug-target interaction.
Thus, graph neural networks have been applied whenever graph structures
are available \cite{do2019attentional,Jiang2020Drug-targetMaps,Nguyen2021GEFA:Prediction}.
More elaborate techniques use self-attention to model the residue-atom
interactions between drug and protein \cite{Zheng2020PredictingSystem,Nguyen2021GEFA:Prediction}.

\subsection{Molecular optimization and generation}

\label{sec:molopt}

The second question is what kind of molecules interact with the given
target. The traditional combinatorial chemistry approach uses a template
as a starting point. From this template, a list of variations is generated
with the goals that they should bind to the pocket with good pharmacodynamic,
have good pharmacokinetics, and be synthetically accessible. The space
of drugs is estimated to be $10^{23}$ to $10^{80}$ substances but
only $10^{8}$ substances have been synthesized thus far. Thus, it
is practically impossible to model this space fully. The current techniques
for graph generations can be search-based, generative, or a combination
of both approaches. The search-based approach starts with the template
and uses optimization framework such as Bayesian Optimization to improve
it over time. This approach does not require a large amount of data
but demands a reliable evaluator through expensive computer simulation
or lab experiments. A more ambitious approach is building expressive
generative models of the entire chemical space, and thus it requires
a large amount of data to train.

\subsubsection{Search-based optimization}

\label{sec:searchopt} The search-based approach can be formulated
as structured machine translation. Here we search for an inverse mapping
of the knowledge base and binding properties back to query molecules.
In this approach, the template molecule is represented as a graph
or a string. The starting molecule is optimized towards desirable
properties. There are two common strategies for optimization. The
first strategy is \emph{sequential optimization in the discrete chemical
space} via atom/bond addition/deletion while maintaining the validity
of the molecule. This sequential discrete search fits well to the
reinforcement learning frameworks with target molecule properties
as rewards \cite{Zhou2019OptimizationLearning}. Reinforcement learning
can cooperate with the graph representation of the molecule with a
graph policy network \cite{You2018GraphGeneration}.

The second strategy is \emph{continuous optimization in the latent
representation space}. We first encode the input molecule graphs or
strings into the latent space. The encoder architecture depends on
the input molecule representation, varying from sequence-based encoder
(e.g., RNN \cite{Gomez-Bombarelli2018AutomaticMolecules} or molecule
graph junction tree \cite{Jin2018JunctionGeneration}). To have an
embedding space representing a set of specific properties, the encoder
is jointly trained with the property prediction task \cite{Gomez-Bombarelli2018AutomaticMolecules}.
Then the molecule is optimized in the latent space with Bayesian Optimization
(BO) \cite{Jin2018JunctionGeneration} or genetic algorithms \cite{Winter2019EfficientSpace}
before being decoded back to the original molecule space. The bottleneck
of this approach is in accurate modeling of the drug latent space.

\subsubsection{Generative molecular generation}

\label{sec:genmol} The molecule optimization can be viewed as inverse
function learning where we learn the function that maps the desired
outputs to the target structure. Then we can leverage the existing
data and query the simulators in an offline manner. In particular,
we start with randomly sampled structure $x$ variable. Then the simulators
answer the query structure $x$ with the properties $y$. With a sufficiently
large number of $\left(x,y\right)$ pairs, the machine learning can
learn the inverse function $x\approx g\left(y\right)$.

The core idea of generative models is learning and sampling from the
density function $p(x)$ of the training data. The main challenges
are due to the complexity of the discrete molecular space, unlike
those typically seen in continuous domains like computer vision. The
most popular generative models to date are variational autoencoder
(VAE), generative adversarial networks (GAN), autoregressive models
and normalizing flow models.

\textbf{VAE} is a two-stage process: we first encode the visible input
structure into the hidden variable and then decode back to the original
structure. The first VAE implemented in modeling the drug space was
by mapping the SMILES sequence into the vector space \cite{Gomez-Bombarelli2018AutomaticMolecules}.
Then we explore the vector space by optimization methods such Bayesian
Optimization (BO) \cite{Gomez-Bombarelli2018AutomaticMolecules} or
genetic algorithms \cite{Winter2019EfficientSpace}. GraphVAE \cite{SimonovskyGraphVAE:Autoencoders}
operates directly on the expressive graph representations. Since the
iterative generation of discrete structure such as graph is non-differentiable,
GraphVAE models the decoded graph as probabilistic fully-connected
on the restricted $k$-node domain. Then the decoded graph is compared
with the ground truth by a standard graph matching. The main drawbacks
of searching in the latent space of VAEs is that it cannot explore
the low density regions, where most interesting novel compounds reside.
A more intrinsically explorative strategy is through compositionality,
where novel combinations can be generated once the compositional rules
are learnt. This has been studied under GrammarVAE \cite{Kusner2017GrammarAutoencoder},
an interesting method that imposes a set of SMILES grammar rules via
parse trees to ensure the validity of the SMILES sequence.

\textbf{GAN} is a powerful alternative to VAEs as it does not require
an encoder, and hence it models the compound distribution \emph{implicitly}.
GAN has two sub-models: a discriminator and a generator. The discriminator
determines if any two samples come from the same distribution. The
generator learns the to generate good samples by trying to fool the
discriminator to believe that the generated samples are real training
data. Mol-CycleGAN \cite{Maziarka2020Mol-CycleGAN:Optimization} learns
the mapping function $G:X\rightarrow Y$ and $G:Y\rightarrow X$ with
two discriminators $D_{X}$ and $D_{Y}$ where $X$ is the set of
input molecule and $Y$ is the molecule set with desired properties.
This ensures the generator transform input molecule to the desired
properties while retaining the structure.

\textbf{Autoregressive models} factorize the density function $p(x)$
as: 
\begin{equation}
p(x)=\prod_{i=1}^{n}p(x_{i}|x_{1},x_{2},...,x_{i-1})
\end{equation}
hence allowing generation of molecules in a stepwise manner. GraphRNN
\cite{You2018GraphRNN:Models} encodes a sequence of graph states
using RNN. Each state represents a step in the graph generation process.
GraphRNN uses BFS to reduce the complexity of learning all the possible
graph state sequences.

\textbf{Normalizing flow models} explicitly learn the complex density
function by transforming the simple distribution through a series
of invertible functions. GraphAF \cite{Shi2020GraphAF:Generation}
defines an invertible function mapping the multivariate Gaussian distribution
to a molecular graph structure. Each step of molecule generation samples
random variables to map them to atom/bond features.

\subsection{Retrosynthesis}

\label{sec:synthesis}

The third question is given a molecule graph, how can we synthesize
the target molecule? The problem is known as retrosynthesis planning,
and it involves determining a chain of reactions to finally synthesize
a target molecule with high efficiency and low cost. At each reaction
step, we need to identify a set of reactants for an intermediate molecule.
This problem can be viewed as the reverse of chemical reaction prediction.
Normally, we have chemical reaction prediction where we predict the
post-reaction products of two or more molecules. However, in retrosynthesis,
given the post-reaction product, the task is to search for two or
more feasible candidates for chemical reaction. Both reaction prediction
and retrosynthesis can be cast as \emph{graph morphism, }where the
molecules form a graph of disconnected sub-graphs, each of which is
a molecule. Reaction changes the graph edges (dropping bonds and creating
new bonds) but keeps the nodes (atoms) intact. A learnable graph morphism
was introduced in GTPN \cite{Do2019GraphPrediction}, a reinforcement
learning based technique to sequentially modify the bonds.

There are two main approaches to solve the retrosynthesis problem:
template-based and template-free. The \textbf{template-based} approach
relies on the set of predefined molecules to construct the target
molecules. This approach formulates the retrosynthesis as the subgraph
matching problem to match the template to the target molecule. The
matching problem is then solved by a variety of techniques, ranging
from a simple deep neural network \cite{Baylon2019EnhancingClassification}
to a more sophisticated framework such as conditional graphical models
\cite{Dai2019RetrosynthesisNetwork}. The template-based approach
suffers from poor generalization on unseen structures as it relies
on predefined fragments and template libraries.

The \textbf{template-free} approach is proposed to overcome the poor
generalization of the template-based approach by inheriting the strong
generalization from the (machine) translation model. The Transformer
model can effectively solve the retrosynthesis problem formulated
as sequence-to-sequence, in which the product molecule SMILES sequence
is translated into a set of reactants SMILES sequences \cite{Karpov2019ARetrosynthesis}.
Graph-to-graph is another approach \cite{Shi2020APrediction}, in
which at first the reaction center is identified using edge embedding
of the graph neural network to break the target molecule into synthons.
Then the synthons are translated into reactants using graph translation
model.

\section{Knowledge-based Reasoning \label{sec:Knowledge-based-Reasoning}}

The biomedical community has accumulated a vast amount of domain knowledge
over the decades, among them those structured as knowledge graphs
are the most useful for learning and reasoning algorithms. We are
primarily interested in the knowledge graphs that represent the relationships
between biomedical entities such as drug, protein, diseases, and symptoms.
Examples of manually curated databases are OMIM \cite{JS2019OMIM.org:Relationships}
and COSMIC \cite{SA2017COSMIC:High-resolution}. Formally a knowledge
graph is a triplet $\mathcal{K}=\left\langle H,R,T\right\rangle $
where $H$ and $T$ are the set of entities, $R$ is the set of relationship
edge connecting entities of $H$ and $T$.

Biomedical knowledge graphs enable multiple graph reasoning problems
for drug discovery. Among them is \emph{drug repurposing}, which aims
to find novel uses of existing approved drugs. This is extremely important
when the demands for new diseases are immediate, such as COVID-19;
when the market is too small to warrant a full \emph{de novo} costly
development cycle (e.g., rare, localized diseases). Given a knowledge
graph, the drug repurposing is searching for new links to a target
from existing drug nodes -- a classic \emph{link prediction} problem.
This setup is also used in \emph{gene-disease prioritisation} in which
we predict the relationship between diseases and molecular entities
(proteins and genes) \cite{Paliwal2020PreclinicalGraphs}. Another
reasoning task is \emph{polypharmacy prediction} of the adverse side
effects due to the interaction of multiple drugs. The multi-relation
graph with graph convolution neural network can encode the drug-drug
interactions \cite{Zitnik2018ModelingNetworks}. Given a pair of drugs,
the drugs are embedded using the encoder and the polypharmacy prediction
task is formulated as a link prediction task.\emph{ }

In what follows, we briefly discuss two major AI/ML problems: \emph{graph
construction} and \emph{graph reasoning}.

\subsection{Automating biomedical knowledge graph construction}

Biomedical knowledge graph is constructed using existing databases
or a rich source of data from biomedical publications. As manual literature
curation is time-consuming, ML has been applied to speed up the process.
The usual framework starts with relevant sentences filtering, followed
by biomedical entity identification and disambiguation \cite{Weber2021HunFlair:Recognition}.
The biomedical entities relationships are extracted from selected
text using rule-based method \cite{Muller2018TextpressoLiterature},
unsupervised \cite{Szklarczyk2021TheSets}, or supervised manner \cite{Li2016BioCreativeExtraction}

\subsection{Reasoning on biomedical knowledge graphs}

Reasoning on knowledge graphs is the process of inferring the relationship
between a pair of entities as well as the logic behind the relationship.
Machine learning reasoning applying to this problem can be categorized
into rule-based reasoning, embedding-based reasoning, and multi-chain
reasoning.

\subsubsection{Rule-based reasoning}

Rules-based reasoning uses logic rules or ontology to infer the new
triplet from the knowledge graph $KG$. A logic rule is defined by
its head $r(x,y)$ and body $B=\left\{ B_{1},B_{2},...,B_{n}\right\} $:
\begin{equation}
r(x,y)\leftarrow B_{1}\land B_{2}\land...\land B_{n}
\end{equation}
AMIE \cite{Galarraga2013AMIE} explores the knowledge graph with the
mining scheme similar to association rule mining. Ontology is the
formal way to describe the types, categories of entities' structure.
Web ontology language (OWL) is a logic-based language to describe
the entities and their relationship. OWL can apply to complex structure
like biomedical knowledge graphs \cite{Chen2013OWLData}.

\subsubsection{Entity and relation embedding}

The logic-based reasoning suffers from the lack of generalization.
A more robust technique assumes a distributed representation of entities
and relations, typically as embedding vectors in high-dimensional
spaces. \textbf{Matrix/tensor factorization} projects the high-dimensional/multi-way
objects into multiple low dimensional vectors. TriModel \cite{Mohamed2020DiscoveringEmbeddings}
learns a low-rank vector representation $\Theta_{E}$ and $\Theta_{R}$
of knowledge entities $\mathbb{E}$ and relations $\mathbb{R}$. The
graph embedding encoder is trained using tensor factorization where
each entity is represented by three embedding vectors. The \textbf{Distance-based
models} exploit the fact that given a triplet $(h,r,t)$, the embedded
representation of $h$ and $t$ is in the proximity, translated by
the embedded vector of the relationship $r$. The best known model
TransE \cite{Bordes2013TranslatingData} learns the embedding of entities
by minimizing the distance between $h+r$ and $t$ and maximizing
the distance between the $h+r$ and $t'$ where $(h,r,t')$ triplet
does not hold.

Structural information like 2D structure or 3D conformation is also
helpful for entity representation learning. It is necessary to integrate
heterogeneous information with structural information. The knowledge
graph embedding can be combined with the structural embedding using
neural factorization machine to form the hybrid representation \cite{Ye2021ASystem}.

\subsubsection{Multi-chain reasoning}

Shallow embedding has achieved remarkable results in reasoning over
the biomedical graphs. However, they can fail to reason when presented
with multiple complex relationships. Multi-chain reasoning extends
the reasoning from a triplet to an extended path of reasoning chain.
DeepPath \cite{Xiong2017DeepPath:Reasoning} applies reinforcement
learning (RL) to find the optimal path of reasoning in the knowledge
graph. The RL can be combined with pre-defined logic rules to learn
the drug repurposing to achieve explainable reasoning \cite{Liu2021NeuralGraphs}.

\section{Challenges and Opportunities \label{sec:Challenges-and-Opportunities}}

\label{sec:challenge} We are now in a position to discuss remaining challenges and chart
possible courses to overcome them.

\paragraph{Large biomedical space}

The drug space is estimated to be from $10^{23}$ up to $10^{60}$.
Due to the diversity of the molecules in term of function and structure,
and the combinatorial nature of their interaction, unconstrained exploration
of the biomedical space--such as molecule optimization and generation--is
intractable. Search-based optimization requires an accurate predicting
model which maps the generated molecule to the target properties.
The generated molecule may have undiscovered properties which leads
to an inaccurate predicting model. As a result, the search direction
can be misleading. Human implicit and explicit feedback can assist
and redirect the optimization to the correct course.

\paragraph{Data quality}

Poor data quality will have a snowball effect in the multi-stage process
of discovery. The data error may lead to an inaccurate machine learning
model when trained on insufficient data. Factors affecting the data
quality are data entry error, hidden bias, and incompleteness due
to law and regulations. Machine learning techniques can help enhance
the data quality by detecting and removing the hidden bias in the
early stage of data collection, data pre-processing, or considering
the bias in the model design. One promising direction is to develop
a ``foundation model'' trained on large-scale data and then adapted
to a wide-range of relevant downstream tasks, similar to what is happening
in the space of text and vision \cite{Bommasani2021OnModels}.

\paragraph{Large gap between virtual screening and real clinical trials}

There is a large gap between clinical trial results and \emph{in silico}
results \cite{Viceconti2021InProducts}: Clinical trials can fail
despite excellent model prediction. For example, machine learning
only predicts the interaction between a drug and a protein without
factoring in a chain reaction or off-target interaction that reduces
the effectiveness of the drug. It is necessary to have a drug discovery
framework that takes account of multiple and chain drug-target, drug-drug,
and protein-protein interactions.

\paragraph{Drug effect on the protein functions}

The current drug discovery and optimization work on the binding interaction
between the target protein and drug molecule. The machine learning
molecule generation and optimization work on the principle of targeting
a specific set of properties or proteins. The machine learning framework
tries to generate or optimize a molecule that is likely to fit to
the binding pocket of the protein. However, there is no clear connection
between the binding activity predicted by the machine learning framework
and the target protein function change. This opens up the direction
to cooperate the protein function information from other sources like
literature into the optimization model.

\paragraph{Personalized prescription and drug discovery}

Personalized medicine allows efficient and safe treatment by coursing
the treatment based on the patient's genomic environments. With the
advance in the 3D printing techniques in pharmaceutics \cite{Goole20163DSystems},
a patient-tailored drug delivery system allows safe and efficient
usage of drugs. At the same time, with the development of bio-markers
in both clinical and biomedical data, the information from bio-markers
is getting integrated into the drug discovery loops. From the machine
learning point of view, it presents a challenge as well as an opportunity
in personalized medicine and drug discovery systems. With the advance
in generative models and optimization, the machine learning framework
can combine bio-maker data with the high-speed drug screening, optimization
and printing techniques to develop a personalized drug discovery system.

\paragraph{Efficient human-machine co-creation}

The end goal of the drug discovery process and the intermediate goal
of machine learning systems may not align due to undiscovered knowledge.
Having an efficient human-machine ecosystem allows the domain experts
to inject prior knowledge, verify and discover the underlying mechanism.

\section{Conclusion}

\label{sec:conclusion} We have provided a survey on recent AI advances targeting one of the
most impactful areas of our time: drug discovery. While this is a
very challenging task, the rewards are huge, and AI is already making
solid progress, contributing to the saving of development costs, and
speed up the discovery. Reversing Eroom's Law will demand new fundamental
advances in AI itself, from learning in the low-data regime, to explore
the vast molecular space, to sophisticated reasoning, to robotic automation.
AI will need to work alongside humans and help expand the knowledge
bases and then benefit from it.

 {\footnotesize{}\bibliographystyle{named}
\bibliography{references}
}{\footnotesize\par}
\end{document}